\documentclass[10pt,journal,compsoc]{IEEEtran}
\ifCLASSOPTIONcompsoc
  \usepackage[nocompress]{cite}
\else
  \usepackage{cite}
\fi
\ifCLASSINFOpdf
   \usepackage[pdftex]{graphicx}
\else
\fi
\usepackage{amsmath, amssymb}
\usepackage{url}

\usepackage{booktabs,multirow}
\usepackage{subcaption}
\begin{document}
%
\title{A Topological Loss Function for \\ Deep-Learning based Image Segmentation \\ using Persistent Homology}
%
%
%
%

\author{James R. Clough,
Nicholas Byrne,
Ilkay Oksuz,
Veronika A. Zimmer,
Julia A. Schnabel,
Andrew P. King
\IEEEcompsocitemizethanks{\IEEEcompsocthanksitem School of Biomedical Engineering and Imaging Sciences, King's College London\protect\\
\IEEEcompsocthanksitem Ilkay Oksuz is also at Istanbul Technical University, Turkey, Computer Engineering Department\protect\\
E-mail: andrew.king@kcl.ac.uk
}
\thanks{Manuscript received XXX; revised XXX.}}

\IEEEtitleabstractindextext{%
\begin{abstract}
We introduce a method for training neural networks to perform image or volume segmentation in which prior knowledge about the topology of the segmented object can be explicitly provided and then incorporated into the training process.
By using the differentiable properties of persistent homology, a concept used in topological data analysis, we can specify the desired topology of segmented objects in terms of their Betti numbers and then drive the proposed segmentations to contain the specified topological features.
Importantly this process does not require any ground-truth labels, just prior knowledge of the topology of the structure being segmented.
We demonstrate our approach in four experiments: one on MNIST image denoising and digit recognition, one on left ventricular myocardium segmentation from magnetic resonance imaging data from the UK Biobank, one on the ACDC public challenge dataset and one on placenta segmentation from 3-D ultrasound.
We find that embedding explicit prior knowledge in neural network segmentation tasks is most beneficial when the segmentation task is especially challenging and that it can be used in either a semi-supervised or post-processing context to extract a useful training gradient from images without pixelwise labels.
\end{abstract}

}

\maketitle

\IEEEdisplaynontitleabstractindextext

%
\IEEEpeerreviewmaketitle

\IEEEraisesectionheading{\section{Introduction}\label{sec:introduction}}
%
%
%
%
%
\IEEEPARstart{S}{egmentation}
is the process of assigning a meaningful label to each pixel in an image and is one of the fundamental tasks in image analysis.
It is required for many applications in which a high-level understanding of the scene, and the presence, sizes, and locations of objects in an image are required, and it is a precursor to many image processing pipelines.
Significant progress has been made on this problem in recent years by using deep convolutional neural networks (CNN), which are now the basis for most newly developed segmentation algorithms \cite{Bernard2018}.
Typically, a CNN is trained to perform image segmentation in a supervised manner using a large number of labelled training cases, i.e. paired examples of images and their corresponding segmentations \cite{litjens2017survey}.
For each case in the training set, the network is trained to minimise some loss function, typically a pixel-wise measure of dissimilarity (such as the cross-entropy) between the predicted and the ground-truth segmentations.
However, errors in some regions of the image may be more significant than others, in terms of the segmented object's interpretation, or for downstream calculation or modelling.
In some cases this can be captured by alternative loss functions, such as the weighted cross-entropy or the generalised Dice loss \cite{sudre2017generalised} which can weight the contribution from rarer classes more strongly.
Nonetheless, loss functions that only measure the degree of overlap between the predicted and the ground-truth segmentations are unable to capture the extent to which the large-scale structure of the predicted segmentation is correct, in terms of its shape or topology.
In principle a large enough training dataset of images and corresponding segmentations will contain enough information for these global features to be learned.
In practice such datasets are rare because ground-truth labels can be expensive to acquire.
They often require a highly trained expert to manually annotate the image, and in the case of segmenting 3D volumes as is frequently required in medical imaging applications, the process can take several hours per volume. 

As reviewed in section \ref{sec:related_work}, there has been significant recent interest in incorporating high-level shape and topological features within CNN training, including the development of specialised segmentation loss functions.
A fundamental obstacle is that this loss function must be differentiable with respect to the class probabilities assigned to each pixel, which is challenging when the presence or absence of particular global features is a discrete quantity.
Here we build on our preliminary work in \cite{Clough2019} to include a loss function for image or volume segmentation which measures the correspondence of the predicted segmentation's topology with that supplied as prior knowledge.
We use the theory of persistent homology (PH), as reviewed in section \ref{sec:theory}, to measure the robustness of the presence of various topological features.
PH allows us to do this in such a manner that a gradient to this loss can be calculated and back-propagated through the weights of the CNN, training it to provide segmentations which are both pixel-wise and topologically accurate.

Our approach is demonstrated in section \ref{sec:experiments} with four experiments.
Firstly, in section \ref{sec:experiment_1} we illustrate the principle of our method by showing that a CNN trained to de-noise MNIST handwritten digits can improve its performance by matching the topology of the digit in question.
We also observe the effect of choosing different topological priors for the same input, showing that the same ambiguous image can be de-noised differently depending on the expected digit's topology.

Then, in section \ref{sec:experiment_2}, we apply the method to the task of segmenting the myocardium of the left ventricle of the heart from short-axis view 2D cardiac magnetic resonance (CMR) images.
By including our topological loss function and the prior knowledge that, from the view in question, the myocardium is ring-shaped, the Dice score of the resulting segmentations is improved, as is their topological accuracy.

In section \ref{sec:experiment_3}, we demonstrate our method on a publicly available challenge dataset, also for the task of segmenting the myocardium from CMR images.

In section \ref{sec:experiment_4} we demonstrate our method on a different imaging modality (ultrasound) and on 3D volumetric data, by applying it to the task of segmenting the placenta.
By incorporating the prior knowledge that the placenta forms one connected component with no topological handles or cavities, we again find that the Dice score of the predicted segmentations improves, as does the topological accuracy of the resulting segmentations.

Finally we discuss other potential applications, generalisations of our method, and other approaches for integrating the power of deep learning with the strong anatomical prior knowledge available to other medical imaging applications.

\section{Related Work}
\label{sec:related_work}
\subsection{Shape constraints in CNN segmentation}
CNNs can be used to perform image segmentation on a pixel-wise basis, with each pixel having some value assigned to it which represents the probability that it is in the segmented object.
These values are not necessarily independent from one pixel to another, as each is determined by weights and activations in the network which will also affect the value given to other pixels.
However, the loss function used to train such networks is typically a function like the binary cross-entropy or Dice score, which measures the overlap between the proposed and the ground-truth segmentations considering each pixel independently.
It can therefore be challenging to train the network to produce segmentations which are coherent in a global sense \cite{Kohl2018}.
Attempts to include some form of global information in the training of such networks have emerged in response to this problem.

In \cite{Mosinska2018} a pre-trained VGG network \cite{simonyan2014very} was used to compare the predicted and the ground-truth segmentations.
The differences in the activations at intermediate layers in this network were used as a secondary loss function (alongside the cross-entropy) which measures the similarity between the two segmentations in a more globally aware manner and was empirically shown to be sensitive to certain topological changes.
However, it is unclear which kinds of high-level features of the segmentations this VGG network measures, and which will be ignored.
A more targeted approach was proposed in \cite{Oktay2018}.
Again, a second neural network was used to compare the proposed and the ground-truth segmentations, but here the second network was an encoder trained on anatomically valid segmentations.
A second loss function was crafted based on the difference between the encoded representations of the two segmentations, and the representation was deliberately designed to efficiently capture features relevant to describing the anatomy in question (which is in this case was also the myocardium as depicted in CMR).
Despite the fact that the encoder was trained on realistic cardiac anatomy, it is hard to know exactly which kinds of shape or topological features are being learned in that approach.
A further limitation of any such methods is that in order to train the network by applying the prior knowledge that the proposed segmentation should be anatomically correct, the ground-truth segmentation is still required, which is not the case in our approach.

An alternative approach to integrating shape priors into network-based segmentation was presented in \cite{Chung2019}.
Here, the segmentation started with a candidate shape which was topologically correct (and approximately correct in terms of its shape), and the network was trained to provide the appropriate deformation to this shape such that it maximally overlapped with the ground truth segmentation.
This work bears similarities to traditional methods using deformable models for segmentation \cite{mcinerney1996deformable} in that an initial shape is deformed to correspond to the image in question, with the important difference that in \cite{Chung2019} the deformation is provided by a neural network rather than found by some energy minimisation procedure.
Such methods can be considered to have a `hard prior' rather than the `soft-prior' of the methods presented above (and by ours) in the sense that the end result can be guaranteed to have the correct shape.
However, this approach may be limited by a requirement that the initial candidate shape be very close to an acceptable answer such that only small shape deformations are needed.
A further potential issue is that the deformation field provided by the network may need to be restricted to prevent the shape from overlapping itself and consequently changing its topology.

\subsection{Neural networks and Persistent Homology}
The differentiable properties of persistent homology \cite{Edelsbrunner2000} make it a promising candidate for the integration of topological information into the training of neural networks.
PH is explained in detail in section \ref{sec:theory}, but the key idea is that it measures the presence of topological features as some threshold or length scale (called the \emph{filtration value}) changes.
Persistent features are those which exist for a wide range of filtration values, and this persistence is differentiable with respect to the original data.
There have recently been a number of approaches suggested for the integration of PH and deep learning, which we briefly review here.

In \cite{Chen2018b} a classification task was considered, and PH was used to regularise the decision boundary.
Typical regularisation of a decision boundary might encourage it to be smooth or to be far from the data.
Here, the boundary was encouraged to be simple from a topological point of view, meaning that topological complexities such as loops and handles in the decision boundary were discouraged.
\cite{Rieck2018} proposed a measure of the complexity of a neural network (considering not just the number of neurons and layers, but also their weights) using PH.
This measure of `neural persistence' was evaluated as a measure of structural complexity at each layer of the network, and was shown to increase during network training as well as being useful as a stopping criterion.

PH has also been suggested as a regularisation on the weights of a neural network, as in \cite{Bruel-Gabrielsson2019}.
There, it was noted that typical regularisation schemes, such as $L_2$ regularisation, effectively stipulate that the network's weights should cluster around a value of $0$.
By using PH on the network's weights, this approach allowed one to instead stipulate, for example, that the weights should form a small number of clusters, but remain agnostic about where those clusters should be.

The topology of learned representations has been considered in \cite{Hofer2019} in which an autoencoder framework was considered, and PH applied to the latent vectors learned by the encoder.
In this way, the representation learned by the encoder can be optimised to respect certain topological, or connectivity properties.
PH has also been used to help train generative adversarial networks in \cite{Charlier2019}.
In this work, the topological properties of the manifolds formed by the real, and generated/fake data were compared in terms of their topology using PH.

In each of these cases, PH was used to measure some set of objects relevant to training neural networks, be it their decision boundaries, weights, activations, learned representations or generated datasets.
The differentiability of the PH measurement is key in that it allows gradient-based optimisation schemes (e.g. stochastic gradient descent) to be used to push the topology of this set of objects towards some desired target.
In some cases, the desired topology is just `as simple as possible'.
In other cases, it is `the same as this other set of objects'.
In others still, it can be specified as some user-defined input, or prior knowledge.
Our method falls into this third category where PH is applied not to the weights or activations of the network, but to the predicted segmentations themselves.

\subsection{Persistent Homology for Image Segmentation}
Some previous literature has applied PH to image segmentation, but the PH calculation has typically been applied to the input image and used as a way to generate features which can then be used by another algorithm.
Applications have included tumour segmentation \cite{Qaiser2016}, cell segmentation \cite{Assaf2017} and cardiac segmentation from computed tomography (CT) imaging \cite{Gao2013}.

The important distinction between these methods and our approach 
is that we apply PH not to the input image being segmented, but rather to the candidate segmentation provided by the network.
To the best of our knowledge, the first work to take this approach was our preliminary work in \cite{Clough2019}, although this idea has subsequently been developed for the specific case of one-dimensional topological features (i.e. connected components) in \cite{Hu2019}. Here, we extend our preliminary work \cite{Clough2019} by introducing an explicit topological loss function which can be used to introduce prior knowledge of any topological feature(s). We also include more extensive experiments on two different medical imaging modalities (including one three-dimensional modality) as well as the MNIST dataset.

By applying PH to the candidate segmentations of a neural network, the topological information found by the PH calculation can be used to provide a training signal to the network, allowing us to compare the topological features present in a proposed segmentation with those specified to exist by some prior knowledge.
Importantly, this can be done even if those topological features are not easily extracted from pixel intensities in the original image.
The mathematical details that describe how PH quantifies the presence of topological features in an image, or a candidate segmentation, are introduced in the next section.

\section{Theory and Methods}
\label{sec:theory}
\subsection{Persistent Homology of Cubical Complexes}

Persistent homology (PH) is a method for calculating the robustness of topological features of a dataset at different scales.
It is part of an emerging field known as topological data analysis, in which ideas from topology are used to extract information from noisy and high-dimensional datasets.
PH has found applications in neuroscience \cite{Sizemore2018}, studying phase transitions \cite{Donato2016}, analysis of tree-structured data \cite{Bendich2016}, and in measuring image artefacts \cite{Han2016}.
Below we give an overview of PH as it applies to our method, but for more thorough reviews we direct the reader to \cite{Edelsbrunner2000, Edelsbrunner2008, Otter2015}.

PH is most often applied to data forming a high-dimensional point cloud, and the topology of simplicial complexes generated from that point cloud is the object of study.
In our case though, the data derives from 2D images or 3D volumes, and so a cubical complex is a more natural representation.
A cubical complex is a set of points, unit line segments, unit squares, cubes, hypercubes, etc.
We define an \emph{elementary interval} as a closed subset of the real line $I=[z,z+1]$ for $z \in \mathbb{Z}$.
Elementary cubes, which will represent pixels or voxels, are the product of elementary intervals, and are given by $Q = I_1 \times I_2 \times ... \times I_k$ where $k$ is the dimension of the space in question.
For simplicity we will describe the two-dimensional case here, and so the region covering the pixel in row $i$ and column $j$ of an image can be denoted by $Q_{ij} = [i, i+1] \times [j, j+1]$.

Consider an $N_x \times N_y$ image represented as a 2D array $\mathbf{X}$, with pixel intensities $X_{ij}$ and a predicted binary segmentation $\mathbf{S}$ also represented as a 2D array with $S_{ij} \in [0,1]$, where $S_{ij}$ is to be thought of as the predicted probability that the pixel in row $i$ and column $j$ of the image belongs to the object being segmented.
$\mathbf{S}$ is calculated by some function $\mathbf{S} = f(\mathbf{X}; \omega)$.
In our case $f$ will be a CNN parameterised by weights $\omega$.
We then consider super-level sets of $\mathbf{S}$, i.e. the set of pixels for which $S_{ij}$ is above some threshold value $p$.
Denoting the super-level sets as $B$:
\begin{align}
     B(p) = \bigcup_{i, j} Q_{ij} : S_{ij} \geq p  
\end{align}
gives us a sequence of sets which grow as the threshold parameter is brought down:
\begin{align}
\varnothing \subseteq B(1) \subseteq B(p_1) \subseteq B(p_2) \subseteq ... \subseteq B(0) \subseteq [0, N_x] \times [0, N_y] .
\end{align}
When $p$ is high, few pixels are in the cubical complex.
As $p$ is lowered, new pixels join the cubical complex and topological features in $B$ are created and destroyed.
Eventually $p=0$ and the entire image is in the super-level set, and so every pixel is in the cubical complex.
PH involves counting the number of topological features of different dimensions in $B(p)$ at each value of $p$, and these numbers are the \emph{Betti numbers} of each cubical complex.
The Betti numbers, $\beta_k$ count the number of features of dimension $k$, where $\beta_0$ is the number of connected components, $\beta_1$ the number of loops or holes, $\beta_2$ the number of hollow voids, etc.
Since our experiments only consider 2D images and 3D volumes, only these first three Betti numbers need to be considered.

\begin{figure}[!t]
\begin{subfigure}{3.25in}
\centering
\includegraphics[width=1.75in]{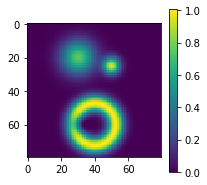}%
\includegraphics[width=1.5in]{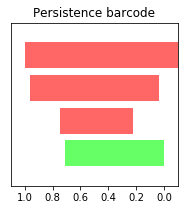}
\caption{Left, an example of a 2D array of size 80x80, which could represent probabilities assigned to each pixel in an 80x80 image.
Right, the barcode diagram of the PH of the super-level sets of this array.
The x-axis of the barcode diagram refers to the filtration value and the ends of each bar correspond to the birth and death filtration values for a particular topological feature.
The ordering of the bars on the y-axis is arbitrary.
Note that the array contains three visible regions of high intensity, which correspond to the three persistent 0-dimensional features shown as red bars in the diagram.
The array also contains a loop of high intensity, corresponding to the one persistent 1-dimensional feature, shown here as a green bar on the barcode diagram.}
\end{subfigure}
\begin{subfigure}{3.25in}
\includegraphics[width=1.75in]{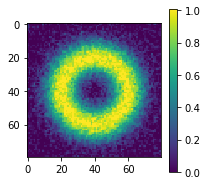}%
\includegraphics[width=1.5in]{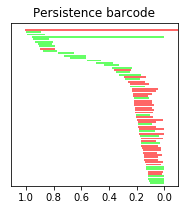}
\caption{Left, a 2D array with a persistent loop feature and additive pixelwise Gaussian noise.
Right, the barcode diagram of the PH of the super-level sets.
The long red and green bars near the top of the barcode correspond to the persistent connected component (red) and loop (green).
The many other smaller bars correspond to topological noise, i.e. the many small loops and connected components which occur only for narrow ranges of the filtration value.}
\end{subfigure}
\caption{Examples of 2D arrays (left) and barcode diagrams describing the persistent homology of their super-level sets (right).}
\label{fig:barcode_diagram}
\end{figure}

The result of this analysis is a set of \emph{birth} and \emph{death} threshold values for each topological feature, which can be represented in a \emph{barcode diagram} like that in Figure \ref{fig:barcode_diagram}.
We will denote the birth and death values for the $\ell$-th longest bar of dimension $k$ as $b_{k, \ell}$ and $d_{k, \ell}$ respectively.
In this diagram $b_{0, \ell}$ is the value at which the $\ell$-th longest red bar begins, and $d_{0, \ell}$ where that bar ends.
Correspondingly $b_{1, \ell}$ is the value at which the $\ell$-th longest green bar begins, and $d_{1, \ell}$ where it ends.
Since we are considering super-level sets $b_{k, \ell} > d_{k, \ell}$.
Those features which have long bars in the barcode diagram (i.e. for which there is a large difference between the birth and death threshold values) are persistent ones which represent meaningful topological features in the data.

As noted in \cite{Clough2019, Bruel-Gabrielsson2019} this calculation of the birth/death values of each feature is differentiable with respect to the values in the image/array - because the values taken by $b_{k, \ell}$ and $d_{k, \ell}$ can only be values in $\mathbf{S}$ - i.e. the birth or death of any feature must occur at the precise value of some particular pixel $S_{ij}$.
This means that for any birth or death threshold value we can calculate its gradient with respect to $\mathbf{S}$, because slightly changing the value of the pixel in question would slightly change the birth or death filtration value of that feature.
Furthermore, since $\mathbf{S} = f(\mathbf{X}; \omega)$, which is also differentiable as $f$ is a neural network, we can calculate the gradient of each $b_{k, \ell}$ and $d_{k, \ell}$ with respect to the network's weights $\omega$.
This will ultimately allow us to adjust the network's weights to make the barcode diagram adhere to our prior knowledge of the topology of the segmented object.

\subsection{Topological Priors}
A differentiable description of the topology of a predicted segmentation allows us to compare that description to prior knowledge about what that topology ought to be, and then use gradient descent to bring it closer to that desired topology.
Let us denote the \emph{desired Betti numbers} of the segmented object by $\beta^*_k$.
Note that in all of our experiments $\beta^*_k$ is prior knowledge determined by the experimenter and is not something that needs to be inferred from the data by an algorithm.
We can then define a loss function for the barcode diagrams as follows:
\begin{align}
\label{eq:loss_function}
 \mathcal{L}_k(\beta^*_k) &= \sum_{\ell=1}^{\beta^*_k} (1 - |b_{k, \ell} - d_{k, \ell}|^2) + \sum_{\ell=\beta^*_k+1}^\infty |b_{k, \ell} - d_{k, \ell}|^2     \\
 \label{eq:loss_function_2}
 \mathcal{L}_{\mathrm{topo}} &= \sum_k \mathcal{L}_k(\beta^*_k)
\end{align}
This loss function is minimised when the barcode diagram has exactly $\beta^*_k$ bars of length $1$, for each $k$ and no other bars%
\footnote{In theory the longest red bar in these diagrams is infinitely long, since for all values of $p$ below zero the entire image is in the cubical complex and so must consist of one connected component only.
In practice we can consider this bar to be cut off at a filtration value of $0$ without affecting any details.}%
.
It is important to note that this loss function does not require knowing the ground truth segmentation, but only the Betti numbers it ought to have.
In the 2D case this is as straightforward as knowing how many connected components and how many loops/holes there are in the object being segmented.
In the case of 3D volumes, the numbers of connected components, loops/handles, and hollow cavities inside the segmented object need to be specified.
Although it is not required for the applications we demonstrate here, we note that it is easily possible to further generalise this framework by changing the summation limits in Equation \ref{eq:loss_function} to allow some values of $\ell$ to not appear in either sum (thereby ignoring the length of some bars in the diagram and so specifying a range of acceptable Betti number values), or by weighting the terms in Equation \ref{eq:loss_function_2} so as to change the relative contribution to the loss from each type of feature. 

\subsection{Implementation}
We will utilise this topological loss function in two frameworks, which we will call the `post-processing framework' and the `semi-supervised framework'.

In the post-processing framework, the segmentation network $f$ is first trained in a supervised manner on a labelled training set of images and pixelwise labels leading to a set of weights $\omega$ which minimise a supervised loss such as the Dice loss, on this training set.
Then, for each item $\mathbf{X}_n$ in the test set (for which the ground-truth segmentation is not available during training but the knowledge of the correct prior topology is), the topological loss function is optimised.
This creates an updated set of network weights $\omega_n$ (which replace $\omega$) for each item  $\mathbf{X}_n$ in the test set, for which the loss function 
\begin{align}
\label{eq:post_proc_framework}
\mathcal{L}(\mathbf{X}_n; \omega, \omega_n) = &
\frac{1}{V}|f(\mathbf{X}_n, \omega) - f(\mathbf{X}_n, \omega_n)|^2 \nonumber \\ 
& + \lambda \mathcal{L}_\mathrm{topo}(\mathbf{X}_n, \omega_n)
\end{align}
is minimised, where $V$ is the number of pixels or voxels in the image or volume.
This effectively finds the minimal change to the output segmentation that corrects its topology.
This framework is appropriate when, for example, each item in the test set has a known topology but these may differ between items.

In the semi-supervised framework the network is trained on a small set of images $\lbrace \mathbf{X}_\ell \rbrace$ and corresponding labels $\lbrace \mathbf{Y}_\ell \rbrace$ and also makes use of a separate larger set of unlabelled images $\lbrace \mathbf{X}_u \rbrace$ whose ground-truth labels are unavailable but whose segmentation topology is known.
For the labelled cases a typical segmentation loss function such as the Dice loss is used, and for the cases which are not labelled the topological loss can be used.
When using this semi-supervised approach in our experiments, we train the network firstly in a fully supervised manner on the small labelled training set before incorporating the unlabelled cases with their topological loss.
This is to ensure that the network's predicted segmentations on the unlabelled cases are sufficiently good that the topological loss can be helpful.
The network is then trained to minimise the loss 
\begin{align}
\label{eq:semisupervised_framework}
\mathcal{L}(\mathbf{X}_\ell, \mathbf{X}_u; \omega) = \sum_{\mathbf{X}_\ell} \mathcal{L}_\mathrm{Dice}(\mathbf{X}, \omega) + \lambda \sum_{\mathbf{X}_u} \mathcal{L}_\mathrm{topo}(\mathbf{X}).
\end{align}
In other words, the total loss is the weighted sum of the normal Dice loss on the labelled cases, and the topological loss, calculated using PH, on the unlabelled cases.
This framework is appropriate when the task is to train one network to segment a set of objects which all have the same topology, and when many images are available but manual annotations may be expensive to acquire.

\section{Experiments}
\label{sec:experiments}
We implemented the CNNs used in our experiments in PyTorch, and the PH calculation using the `TopLayer' Python package introduced in \cite{Bruel-Gabrielsson2019} which uses PyTorch to calculate the PH of images in such a way as to retain the gradients of the birth and death filtration values $b_{k,\ell}$ and $d_{k, \ell}$.
We use the Python module Gudhi \cite{gudhi} to produce the barcode diagrams.
\subsection{Experiment 1}
\label{sec:experiment_1}
\begin{figure}[!t]
\centering
\includegraphics[width=3.5in]{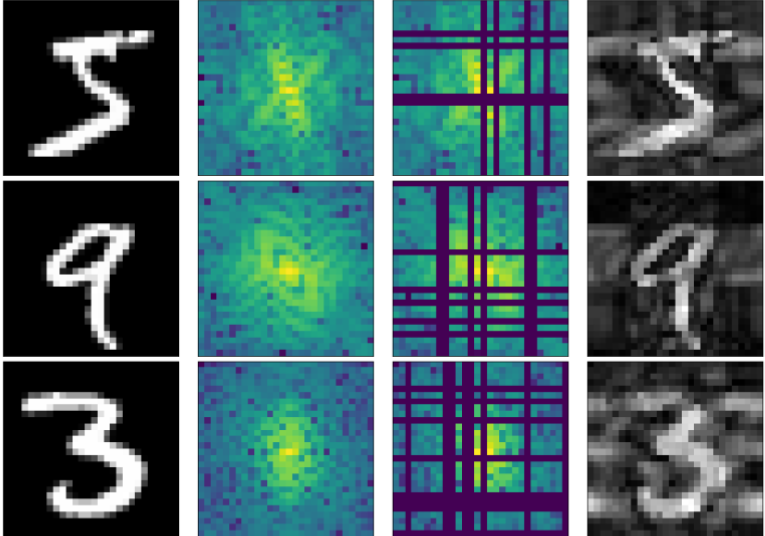}
\caption{Corrupted versions of MNIST digits. Left column, the original images. 
Second column, their Fourier transforms, showing the image in the frequency domain. 
Third column, the Fourier transforms with $m$ horizontal and vertical lines randomly selected and zero-filled.
Right column, the inverse Fourier transform of the third column, showing the original images with artefacts.
Top row: $m=4$, middle row: $m=6$, bottom row: $m=8$.
As $m$ increases the image-domain artefacts are more severe.
Note that since the removal of lines in the frequency domain is random and so not necessarily symmetric in the Fourier domain the images resulting from the inverse Fourier transform are complex-valued.
We take the magnitude only, and then normalise the images to have intensities between 0 and 1.
}
\label{fig:mnist_noise_example}
\end{figure}

\begin{figure}[!t]
\centering
\includegraphics[width=3.5in]{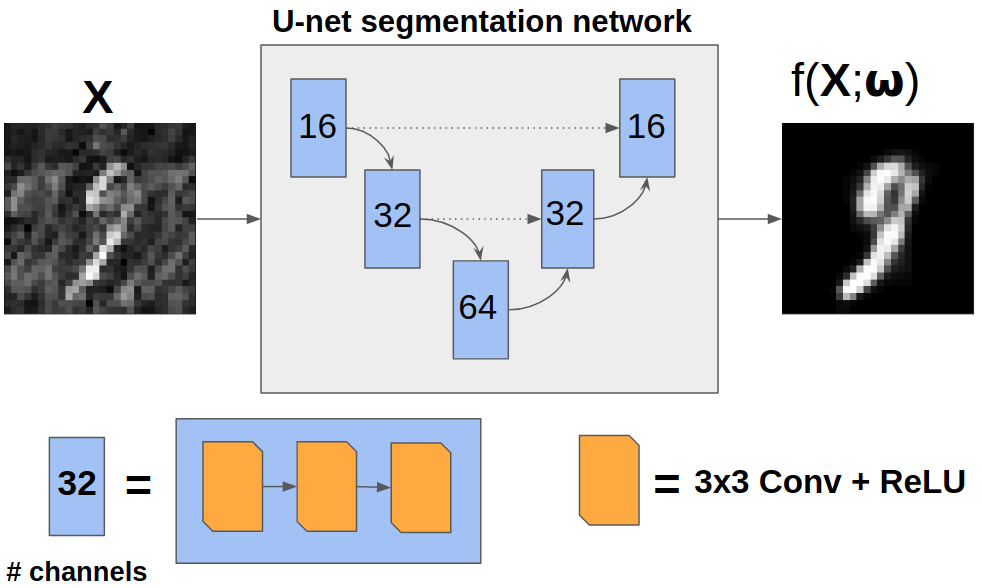}
\caption{Diagram of the simple U-net architecture used in experiment 1. 
Dotted arrows correspond to feature concatenation.
}
\label{fig:mnist_experiment_diagram}
\end{figure}

To illustrate the principle of using topological priors on the image-domain output of a neural network, we demonstrate our approach on a toy experiment: de-noising images from the MNIST handwritten digits dataset \cite{lecun1998mnist}.
We begin by generating corrupted versions of each MNIST digit.
To generate noise with spatial correlations, we corrupt the images by taking the Fourier transform, randomly remove $m$ vertical and $m$ horizontal lines, replace the removed values with zeros, and then take the inverse Fourier transform.
We take the absolute value of the result and renormalise each image to the $[0,1]$ range.
As shown in Figure \ref{fig:mnist_noise_example}, the resulting corrupted images contain various artefacts including blurring and aliasing.
We will denote the corrupted image as $\mathbf{X}$ and the original image as $\mathbf{Y}$.
A simple U-net-like \cite{Ronneberger2015} CNN%
\footnote{Although U-net architectures are more commonly used for segmentation than for image de-noising we use it here for the consistency of adopting a single type of CNN architecture throughout.
The aim in this experiment is to demonstrate our approach rather than produce state-of-the-art results in image de-noising.}
is trained to recover $\mathbf{Y}$ from the corrupted version $\mathbf{X}$, minimising the mean squared error between $f(\mathbf{X}; \omega)$ and $\mathbf{Y}$ as illustrated in Figure \ref{fig:mnist_experiment_diagram}.
We then assess whether adding a further loss function (like that in Equation \ref{eq:loss_function}) in the post-processing framework results in better quality recovered images.


\begin{figure*}[!tb]
\begin{subfigure}{1.2in}
\centering
\includegraphics[width=1.2in]{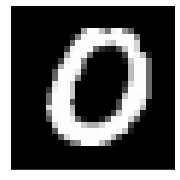}
\vspace{-3mm}
\caption{Original digit.}
\end{subfigure}
\begin{subfigure}{1.2in}
\includegraphics[width=1.2in]{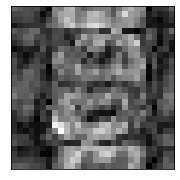}
\vspace{-3mm}
\caption{Corrupted digit.}
\end{subfigure}
\begin{subfigure}{1.2in}
\includegraphics[width=1.2in]{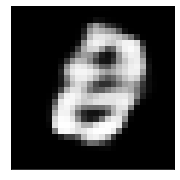}
\vspace{-3mm}
\caption{$\mathcal{L}_\mathrm{MSE}$ output.}
\end{subfigure}
\begin{subfigure}{1.25in}
\vspace{-4mm}
\includegraphics[width=1.25in]{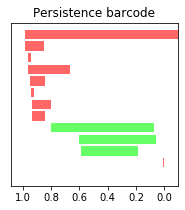}
\vspace{-4mm}
\caption{$\mathcal{L}_\mathrm{MSE}$ barcode.}
\end{subfigure}
\begin{subfigure}{1.2in}
\includegraphics[width=1.2in]{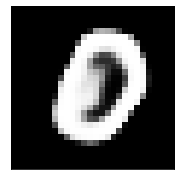}
\vspace{-3mm}
\caption{$\mathcal{L}_\mathrm{topo}$ output.}
\end{subfigure}%
\begin{subfigure}{1.25in}
\vspace{-4mm}
\includegraphics[width=1.25in]{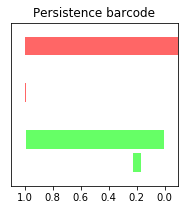}
\vspace{-4mm}
\caption{$\mathcal{L}_\mathrm{topo}$ barcode.}
\end{subfigure}
\caption{This digit a `0', shown in \emph{(a)} should consist of one connected component with one loop, corresponding to one long red bar and one long green bar in the barcode diagram.
The network is given as an input the highly corrupted version of this digit, shown in \emph{(b)}.
The digit reconstructed by the original network, \emph{(c)}, is misclassified as an `8'.
Its barcode diagram, \emph{(d)} has three green bars: an incorrect topology for a `0'.
After applying the topological prior to the reconstruction, the network output \emph{(e)} is correctly classified as a `0'.
Its barcode diagram, \emph{(f)} shows the correct topological features of a `0' digit.}
\label{fig:mnist_reconstruction_example_1}
\end{figure*}
\begin{figure*}[!tb]
\begin{subfigure}{1.2in}
\centering
\includegraphics[width=1.2in]{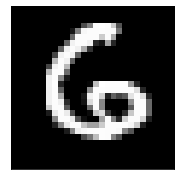}
\vspace{-3mm}
\caption{Original digit.}
\end{subfigure}
\begin{subfigure}{1.2in}
\includegraphics[width=1.2in]{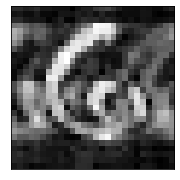}
\vspace{-3mm}
\caption{Corrupted digit.}
\end{subfigure}
\begin{subfigure}{1.2in}
\includegraphics[width=1.2in]{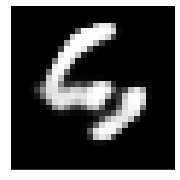}
\vspace{-3mm}
\caption{$\mathcal{L}_\mathrm{MSE}$ output.}
\end{subfigure}
\begin{subfigure}{1.25in}
\vspace{-4mm}
\includegraphics[width=1.25in]{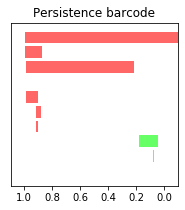}
\vspace{-4mm}
\caption{$\mathcal{L}_\mathrm{MSE}$ barcode.}
\end{subfigure}
\begin{subfigure}{1.2in}
\includegraphics[width=1.2in]{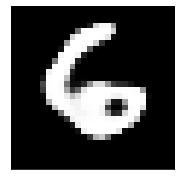}
\vspace{-3mm}
\caption{$\mathcal{L}_\mathrm{topo}$ output.}
\end{subfigure}%
\begin{subfigure}{1.25in}
\vspace{-4mm}
\includegraphics[width=1.25in]{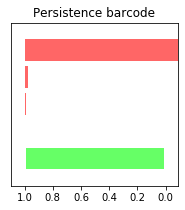}
\vspace{-4mm}
\caption{$\mathcal{L}_\mathrm{topo}$ barcode.}
\end{subfigure}
\caption{This digit a `6', shown in \emph{(a)} should consist of one connected component with one loop, corresponding to one long red bar and one long green bar in the barcode diagram.
The network is given as an input the highly corrupted version of this digit, shown in \emph{(b)}.
The digit reconstructed by the original network, \emph{(c)}, is misclassified as an `5'.
Its barcode diagram, \emph{(d)} has no long green bars: an incorrect topology for a `6'.
After applying the topological prior to the reconstruction, the network output \emph{(e)} is correctly classified as a `6'.
Its barcode diagram, \emph{(f)} shows the correct topological features of a `6' digit.}
\label{fig:mnist_reconstruction_example_2}
\end{figure*}
We first train the CNN in a supervised manner using $N_\ell=100$ paired cases of corrupted and ground-truth digits with a mean squared error loss.
Then, for each digit in the test set (of size $N_\mathrm{test}=1000$) we apply post-processing consisting of minimising a loss function like that in Equation \ref{eq:post_proc_framework} with $\lambda=0.02$.
Recall that $\beta_0^*$ corresponds to the desired number of connected components and $\beta_1^*$ the desired number of holes in the object.
For this experiment we assumed that the digits 1, 2, 3, 4, 5 and 7 have $\beta_0^*=1, \beta_1^*=0$, the digits 6, 9 and 0 have $\beta_0^*=1, \beta_1^*=1$ and the digit 8 has $\beta_0^*=1, \beta_1^*=2$.
We then compare the quality of the reconstructed digits before and after this topologically informed post-processing step.
We assess the quality of the reconstructed digits by computing the mean squared error between the ground truth and the reconstructed digit, but also by measuring how recognisable the resulting digit was.
This is quantified by first training another CNN (the `classifier network') to classify MNIST digits (trained on 10000 digits not used in the main experiment), and assessing how well this network could classify the reconstructed digits.
On the original uncorrupted MNIST digits this classifier network has a classification accuracy of $98.7\%$.
If the reconstructed digits are sufficiently similar to their originals, then this network should be able to classify them with a similar accuracy.

\begin{figure}[!t]
\centering
\includegraphics[width=3.6in]{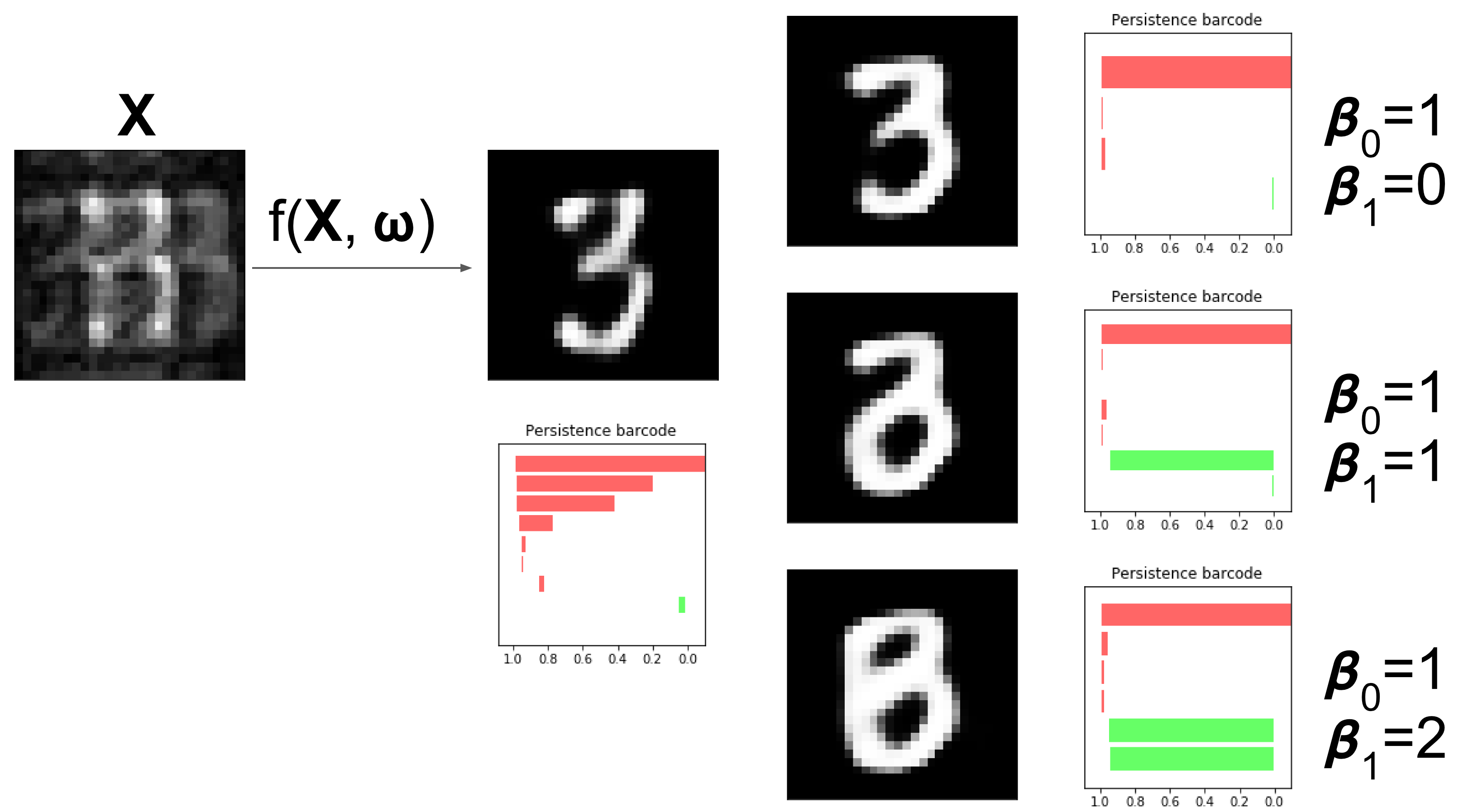}
\caption{The same degraded image is reconstructed in three different ways depending on the topological prior used.
On the left, the corrupted image of a `3' digit, $\mathbf{X}$ is reconstructed by the original network $f(\mathbf{X}; \omega)$.
On the right, three different topological priors are applied for post-processing, each resulting in a modified set of weights $\omega'$ and modified reconstructed digits $f(\mathbf{X}; \omega')$.
The resulting reconstructions have the desired topology in each case.
However they do not necessarily look like a real digit, since topology alone, being invariant to rotations and reflections, is not enough to correctly describe the shape of a digit.}
\label{fig:mnist_different_priors}
\end{figure}

As shown in Table \ref{tab:mnist_results}, the inclusion of the topological prior significantly improved the performance of the classifier network on the reconstructed digits, indicating that the changes in the digit reconstruction (see some examples in Figures \ref{fig:mnist_reconstruction_example_1} and \ref{fig:mnist_reconstruction_example_2}) make them easier to correctly recognise.
\begin{table}[tb]
    \centering
\begin{tabular}{l |c | c | c | c}
& \multicolumn{2}{c}{Classification accuracy}  & \multicolumn{2}{c}{Mean Squared Error}\\
  & $\mathcal{L}_\mathrm{MSE}$ & $\mathcal{L}_\mathrm{topo}$ & $\mathcal{L}_\mathrm{MSE}$ & $\mathcal{L}_\mathrm{topo}$ \\
 \toprule
$m=2$ &  $95.7\%$ & $96.2\%$ & 0.008 & 0.008 \\ 
$m=4$ &  $92.9\%$ & $93.7\%$ & 0.014 & 0.013 \\ 
$m=6$ &  $87.6\%$ & $90.8\%^*$ & 0.023 & 0.023 \\ 
$m=8$ &  $80.4\%$ & $86.0\%^*$ & 0.037 & 0.028 \\ 
$m=10$ & $70.4\%$ & $75.0\%^*$ & 0.032 & 0.033 \\ 
$m=12$ & $56.9\%$ & $63.0\%^*$ & 0.043 & 0.042 \\ 
\end{tabular}
    \caption{Table of results for MNIST experiments.
    As $m$, the number of removed lines in the corrupted images, increases, the classification accuracy falls and mean squared error on the reconstructed digits increases.
    The inclusion of the topological post-processing leads to more coherent reconstructed digits which are more easily classified. $^*$ indicates statistical significance at 95\% confidence with McNemar's test between the classifiers trained using images reconstructed using the $\mathcal{L}_\mathrm{MSE}$ and $\mathcal{L}_\mathrm{topo}$ losses. None of the Mean Squared Error values were statistically significantly different with a 2-tailed Wilcoxon signed rank test at 95\% confidence.
    }
    \label{tab:mnist_results}
\end{table}
In Figure \ref{fig:mnist_different_priors} we show that the same image is de-noised differently depending on the applied topological prior.
In each case the desired topology, as quantified by the barcode diagrams, is reached, resulting in a different looking reconstructed image.

\subsection{Experiment 2}
\label{sec:experiment_2}
\begin{figure}[!t]
\centering
\includegraphics[width=1.6in]{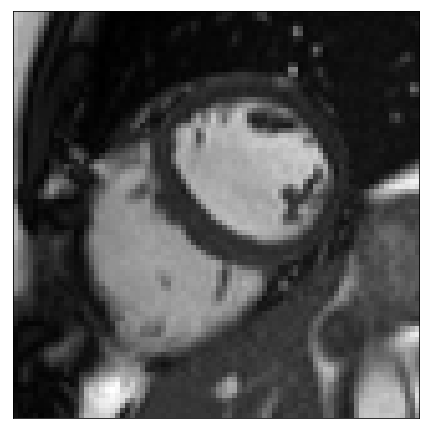}
\includegraphics[width=1.6in]{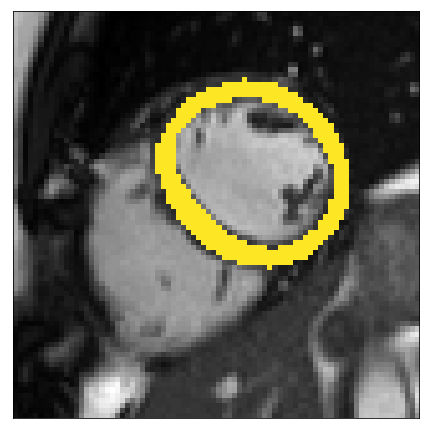}
\includegraphics[width=1.6in]{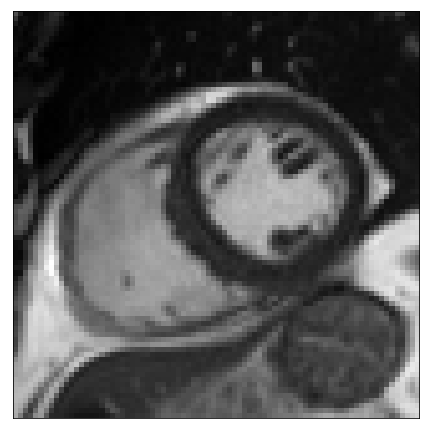}
\includegraphics[width=1.6in]{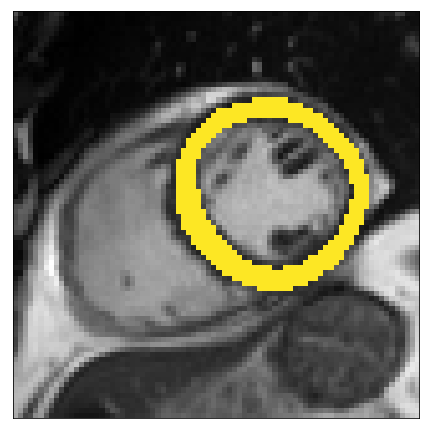}
\caption{Two example short-axis CMR images from the UK Biobank dataset (left) and with manually annotated segmentations of the myocardium (right).}
\label{fig:lv_data_example}
\end{figure}

The second experiment considers the task of segmenting the myocardium of the left ventricle of the heart in CMR images. The data used here are from the UK Biobank \cite{Petersen2016}, and consist of 2D images from the short-axis view of the heart, where we take only the mid-slice of the short-axis stack, from the first cardiac phase from each subject.
This ensures that each image comes from a different subject, and contains approximately the same anatomy, at the same point in the cardiac cycle. 
Each image was cropped to an 80x80 pixel square centred on the left ventricle.
Examples of typical images and manual segmentations from this dataset are shown in Figure \ref{fig:lv_data_example}.

\begin{figure}[!t]
\centering
\includegraphics[width=0.8in]{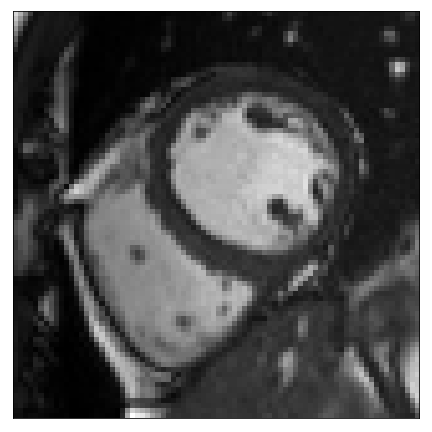}
\includegraphics[width=0.8in]{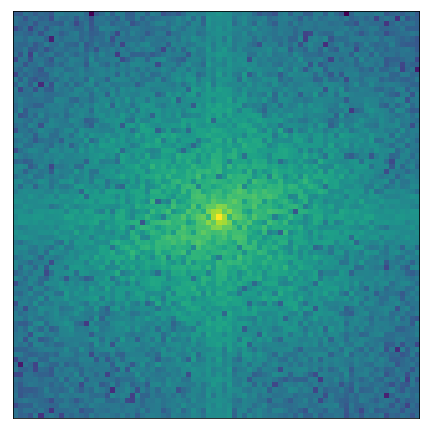}
\includegraphics[width=0.8in]{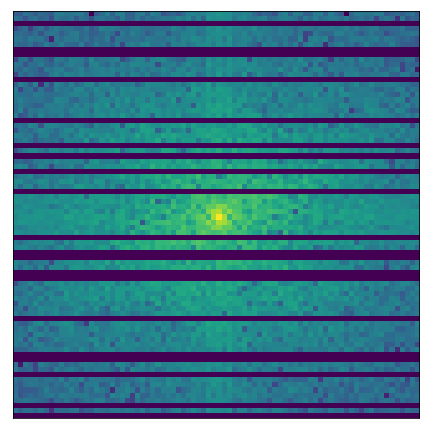}
\includegraphics[width=0.8in]{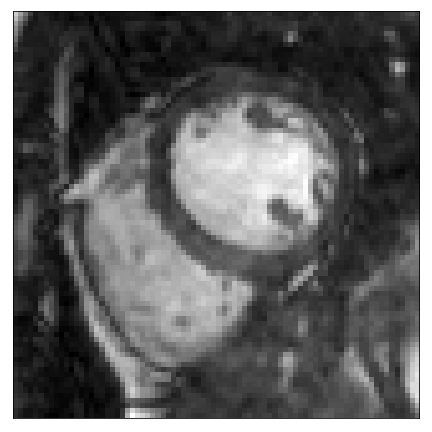}\\ 
\includegraphics[width=0.8in]{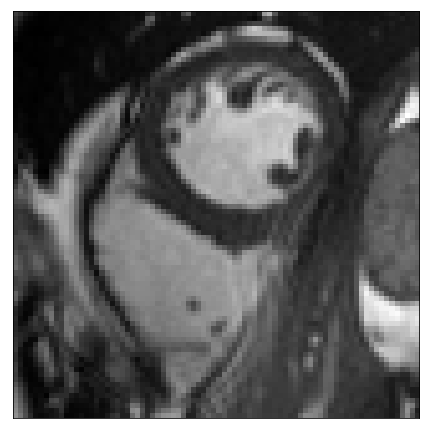}
\includegraphics[width=0.8in]{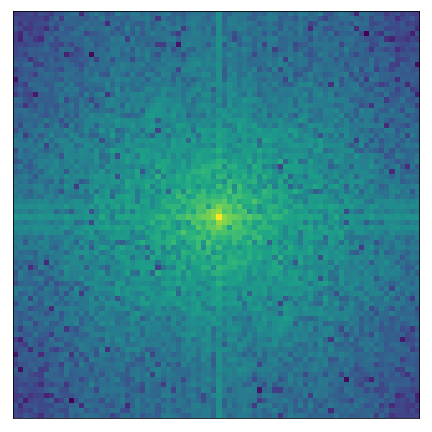}
\includegraphics[width=0.8in]{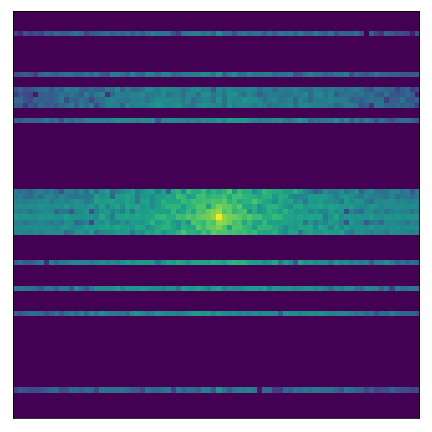}
\includegraphics[width=0.8in]{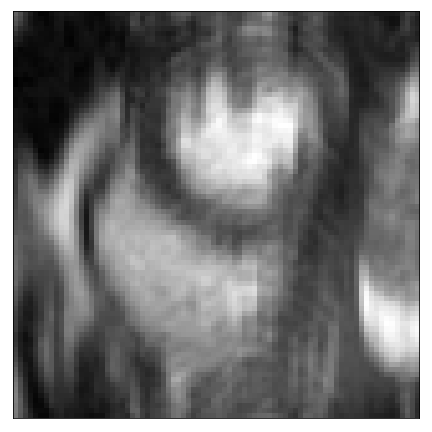}
\caption{Two CMR images artificially degraded by removing lines in the Fourier transform.
From left to right, the original CMR image, the Fourier transform, the degraded Fourier transform, and the inverse Fourier transform of the degraded frequencies.
On the top row, 20 of the 80 frequency lines are set to zero, causing mild image degradation.
On the bottom row, 60 of the 80 frequency lines are set to zero, causing serious image degradation.
In all cases, the middle 8 lines are reserved from deletion.
This process allows us to assess the efficacy of the segmentation CNN on tasks of varying difficulty since segmenting the more strongly corrupted images is a more challenging task.}
\label{fig:lv_data_degraded}
\end{figure}

In this experiment we utilise our method in a semi-supervised framework since a large number of images are available and the desired topology for each segmentation is the same: the myocardium of the left ventricle from the short-axis view is topologically circular.
We expect to see a segmentation which has one connected component with one hole/loop, and so $\beta_0^*=1, \beta_1^*=1$.
In order to assess the utility of our method under a variety of conditions we conduct experiments in which the quality of the images provided as input to the network is degraded to varying degrees.
We do this by introducing artefacts into the data by randomly removing $m$ lines in the Fourier transform of each image and zero-filling them, as shown in Figure \ref{fig:lv_data_degraded}.
The parameter $m$ quantifies the degree to which the image is corrupted, and so as $m$ increases the segmentation task becomes more challenging.
We chose to corrupt the images in this way as corruption during CMR image acquisition occurs in the Fourier domain.

We began by training the segmentation network on a small number of labelled cases, $N_\ell$.
Our approach is compatible with any choice of network architecture, and the focus of our work is to introduce the topological loss function for segmentation, and not to assess the various CNN architectures that have been proposed in the literature.
For the segmentation network we therefore choose a U-net \cite{Ronneberger2015} since it is amongst the most frequently deployed.
We began training in a supervised manner, using a batch size equal to $N_\ell$ and training for up to 3000 epochs.
Since the training sets are in some experiments very small, we mitigated the risk of over-fitting by stopping training early.
A separate validation set of 100 cases was tested every 50 epochs and training stopped early if the Dice score on this validation set did not improve for 5 such tests. 

The topological prior was then introduced by training in a semi-supervised manner with an additional $N_u$ unlabelled images.
Training with the topological prior in a semi-supervised manner consisted of alternating steps of processing a batch of labelled cases and back-propagating the Dice loss through the network, and then processing a batch of unlabelled cases, calculating the PH of their predicted segmentations and back-propagating their topological loss through the network.
The relevant loss function being minimised is that in Equation \ref{eq:semisupervised_framework}, where $\lambda=0.01$.
In this way we can use the large number of unlabelled cases to generate a useful training signal by leveraging the fact that the topology of their segmentations is known, even if those ground truth segmentations are not available.

As a baseline method, we evaluated the performance of the same network architecture using solely supervised training on the small number of labelled cases. We also evaluated the baseline method with the addition of two different postprocessing techniques: morphological closure and the use of a conditional random field technique (CRF) \cite{Krahenbuhl2011}. Both of these approaches can help to correct some topological errors such as small gaps in the segmentation.
The morphological closure operation used a disk-shaped kernel of radius 7 pixels.
For the CRF
approach we used the method described in \cite{Krahenbuhl2011} with parameter settings which were tuned using a grid search to optimise performance on a CMR segmentation task \cite{Bai2017}.
We also compared our method with a boot-strapping semi-supervised approach, similar to that described in \cite{Bai2017} (but without the CRF postprocessing), in which predicted segmentations on the unlabelled cases are used to train the network in an iterative process.
This semi-supervised approach uses the same set of unlabelled cases as does our semi-supervised method with the topological loss.
To assess the output segmentations, we calculated the Dice score with respect to the ground truth, and also counted the proportion of test cases for which the predicted segmentation was topologically correct when thresholded at $p=0.5$.

\begin{figure}[!t]
\centering
\begin{subfigure}{3.2in}
\includegraphics[width=1.6in]{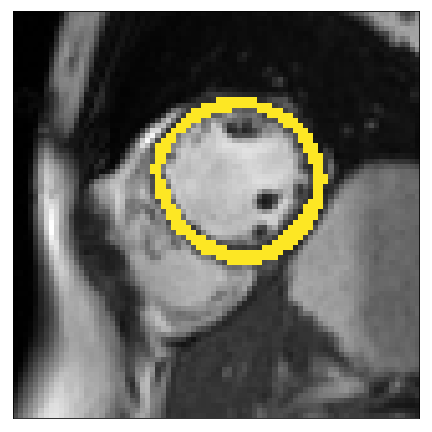}%
\includegraphics[width=1.6in]{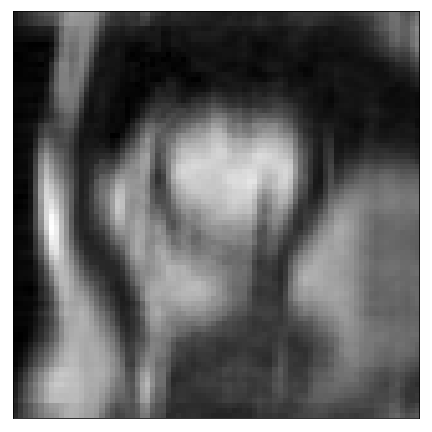}
\caption{Left: Uncorrupted image and ground-truth segmentation.
Right: Corrupted image, the input to the network.}
\end{subfigure}
\begin{subfigure}{3.2in}
\includegraphics[width=1.6in]{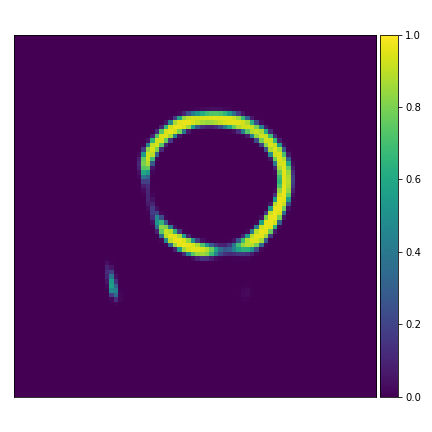}%
\includegraphics[width=1.6in]{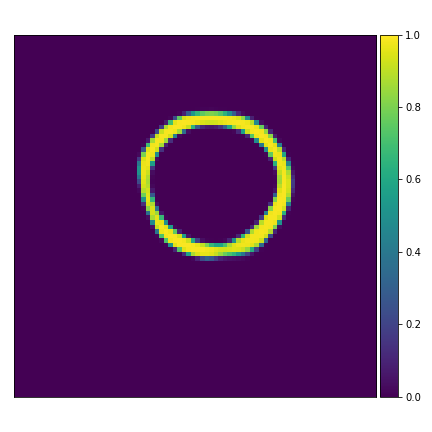}
\caption{Left: The predicted segmentation from the network trained only with supervised learning.
Right: The predicted segmentation from the network trained in a semi-supervised manner, incorporating the topological prior.}
\end{subfigure}
\begin{subfigure}{3.2in}
\includegraphics[width=1.6in]{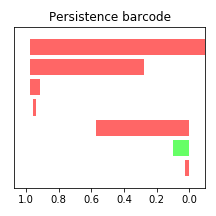}%
\includegraphics[width=1.6in]{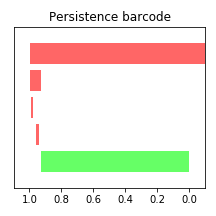}
\caption{Left: The persistence barcode for the predicted segmentation from the network trained only with supervised learning.
Right: The persistence barcode for the predicted segmentation from the network trained in a semi-supervised manner, incorporating the topological prior.}
\end{subfigure}
\caption{Segmentations and barcodes with and without the topological prior.}
\label{fig:lv_topoprior_examples_1}
\end{figure}

\begin{figure}[!t]
\centering
\begin{subfigure}{3.2in}
\includegraphics[width=1.6in]{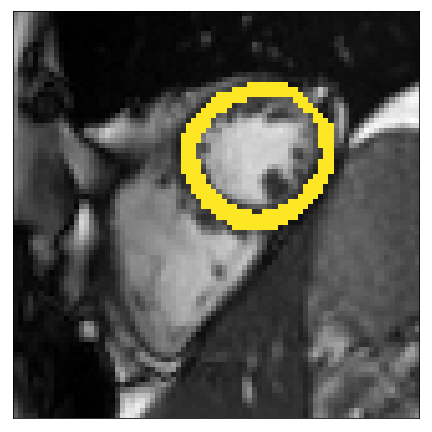}%
\includegraphics[width=1.6in]{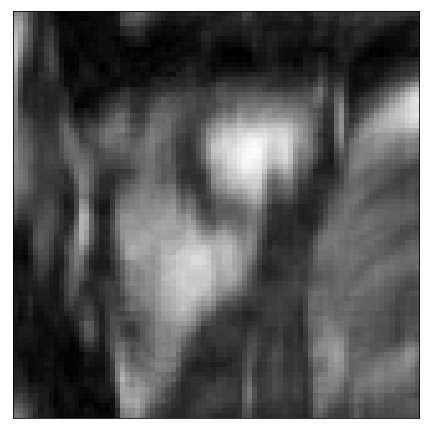}
\caption{Left: Uncorrupted image and ground-truth segmentation.
Right: Corrupted image, the input to the network.}
\end{subfigure}
\begin{subfigure}{3.2in}
\includegraphics[width=1.6in]{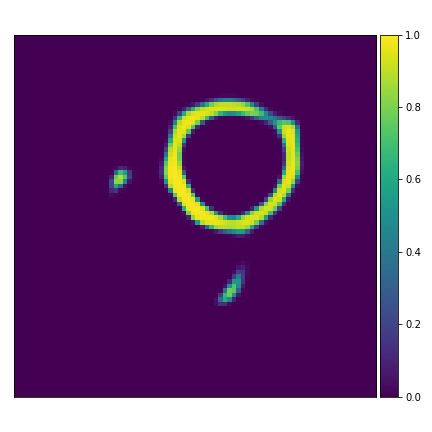}%
\includegraphics[width=1.6in]{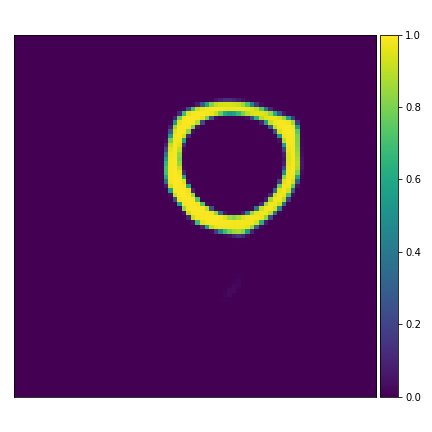}
\caption{Left: The predicted segmentation from the network trained only with supervised learning.
Right: The predicted segmentation from the network trained in a semi-supervised manner, incorporating the topological prior.}
\end{subfigure}
\begin{subfigure}{3.2in}
\includegraphics[width=1.6in]{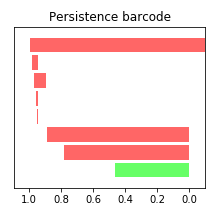}%
\includegraphics[width=1.6in]{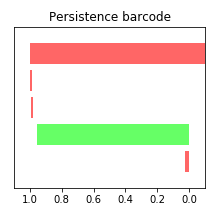}
\caption{Left: The persistence barcode for the predicted segmentation from the network trained only with supervised learning.
Right: The persistence barcode for the predicted segmentation from the network trained in a semi-supervised manner, incorporating the topological prior.}
\end{subfigure}
\caption{Segmentations and barcodes with and without the topological prior.}
\label{fig:lv_topoprior_examples_2}
\end{figure}

Figures \ref{fig:lv_topoprior_examples_1} and \ref{fig:lv_topoprior_examples_2} show typical cases where the network trained using only labelled cases makes a topological error, either segmenting extra connected components, or leaving a gap in the ring-shaped myocardium.
This kind of error is clear in the persistence barcode diagrams in which the extra components appear as additional red bars, and gaps in the myocardium appear as a shortening of the green bar (since the loop feature only appears when the filtration value is brought very close to 0).
After applying our topological prior in training the persistence barcodes are much closer to that which is specified by prior knowledge and which minimises the loss function in Equation \ref{eq:semisupervised_framework}, demonstrating that these topological errors are removed.
Table \ref{tab:lv_results} shows that as well as correcting almost all of these topological errors, the Dice score also generally improves when using the topological loss.
This effect is most significant when the initial segmentation task is challenging (i.e. the images heavily degraded).
This demonstrates that extracting the relevant topological information from the unlabelled images in some way regularises the CNN allowing for better test-set performance even when few manually annotated images are available for training.
Neither the morphology based nor the CRF based postprocessing techniques were able to achieve comparable levels of performance with regard to topological correctness.

Note that the images corrupted by removing \emph{m=60} lines from the Fourier transform were highly corrupted (see Figure \ref{fig:lv_data_degraded}), with 60 out of 80 Fourier lines removed and zero-filled, resulting in a median signal-to-noise ratio of 3.3dB compared to 4.22dB for the original images, and a mean absolute gradient magnitude of 0.0445 compared to 0.0558 for the original images. Therefore, we demonstrate in this experiment that our PH based method is robust within (and beyond) a range of clinically realistic corruption levels.

\begin{table*}[tb]
    \centering
\begin{tabular}{l |c | c | c | c || c | c | c | c |}
& \multicolumn{4}{c}{$m=20$} & \multicolumn{4}{c}{$m=60$} \\
\toprule
& $N_\ell=10$ & $N_\ell=20$ & $N_\ell=40$ & $N_\ell=100$
& $N_\ell=10$ & $N_\ell=20$ & $N_\ell=40$ & $N_\ell=100$\\
 \toprule
 \multirow{2}{*}{Supervised}
& $67.9^* \pm 2.5$ &  $76.3^* \pm 0.7$ & $82.1^* \pm 0.5$ & $86.1 \pm 0.2$
& $62.0^* \pm 2.0$ &  $70.8^* \pm 0.7$ & $76.9^* \pm 0.5$ & $82.1 \pm 0.3$ \\
& $ 42.1^* \pm 5.7 \%$ & $59.8^* \pm 3.1\%$ & $74.0^* \pm 2.2\%$ & $85.9^* \pm 1.0\%$  
& $ 25.8^* \pm 2.9 \%$ & $58.3^* \pm 5.3\%$ & $67.6^* \pm 2.9\%$ & $80.8^* \pm 2.1\%$
\\
 \midrule
Supervised & $68.2^* \pm 2.3$ & $76.2^* \pm 0.6$ & $81.7^* \pm 0.5$ & $85.7^* \pm 0.2$  
           & $62.9^* \pm 1.9$ & $70.8^* \pm 0.6$ & $76.9^* \pm 0.47$ & $82.0 \pm 0.3$   \\
 + closure & $53.8^* \pm 5.2\%$ & $71.3^* \pm 3.2 \%$ & $81.4^* \pm 1.4\%$ & $91.8^* \pm 0.8\%$  
           & $44.4^* \pm 4.7\%$ & $66.9^* \pm 5.0 \%$ & $76.9^* \pm 2.2\%$ & $87.1^* \pm 1.7\%$
 \\
 \midrule
Supervised & $67.4^* \pm 3.8$ & $76.1^* \pm 0.7$ & $81.9^* \pm 0.5$ & $86.0^* \pm 0.2$  
           & $61.7^* \pm 1.5$ & $70.4^* \pm 0.8$ & $76.4^* \pm 0.6$ & $81.9^* \pm 0.4$   \\
 + CRF     & $38.1^* \pm 6.3\%$ & $52.9^* \pm 3.0 \%$ & $69.7^* \pm 2.6\%$ & $82.1^* \pm 1.0\%$  
           & $22.0^* \pm 3.3\%$ & $52.0^* \pm 6.0 \%$ & $61.3^* \pm 3.7\%$ & $77.8^* \pm 2.4\%$
 \\
 \midrule
Semi- & $74.7 \pm 0.6$ &  $78.6 \pm 0.4$ & $83.6 \pm 0.4$ & $87.2 \pm 0.3$  
      & $67.3 \pm 1.3$ &  $73.3 \pm 0.9$ & $78.2 \pm 0.5$ & $82.7 \pm 0.3$ \\
supervised & $57.4^* \pm 4.8 \%$ & $68.1^* \pm 2.3\%$ & $80.8^* \pm 1.7\%$ & $89.8^* \pm 1.4\%$  
           & $48.4^* \pm 6.0 \%$ & $60.9^* \pm 2.6\%$ & $71.3^* \pm 2.8\%$ & $84.5^* \pm 1.4\%$
\\
 \midrule
\multirow{2}{*}{Ours} & $74.3 \pm 0.8$ &  $79.1 \pm 0.4$ & $83.5 \pm 0.3$ & $86.8 \pm 0.2$
                      & $68.6 \pm 0.8$ &  $74.3 \pm 0.6$ & $78.6 \pm 0.4$ & $82.6 \pm 0.3$\\
 & $67.1 \pm 2.9\%$ & $80.5 \pm 2.3 \%$ & $88.1 \pm 1.5\%$ & $93.6 \pm 1.3\%$
 & $62.0 \pm 5.1\%$ & $75.6 \pm 2.4 \%$ & $85.7 \pm 1.8\%$ & $91.2 \pm 1.6\%$
 \\
 \midrule
\end{tabular}
    \caption{Table of results for LV segmentation experiments.
    In this experiment the number of labelled cases was $N_\ell$ and the number of unlabelled cases $N_u=400$.
    The number of lines removed from the Fourier domain data, representing the difficulty of the segmentation task was $m=20$ (left) and $m=60$ (right).
    For each method and experiment, the average Dice score between the predicted segmentation and the ground truth (top), and proportion of the test set which was segmented without topological errors (bottom) is shown.
    For both of these metrics higher scores are better.
    The ranges indicate the standard error over 10 experiments each with different training, validation and test sets. $^*$ indicates statistical significance at 95\% confidence with a 2-tailed Wilcoxon signed rank test between `Ours' and each other evaluated method. }
    \label{tab:lv_results}
\end{table*}

\subsection{Experiment 3}

\label{sec:experiment_3}
To enable comparison with other segmentation models, we also evaluated our method on the publicly available ACDC dataset \cite{Bernard2018}. This dataset includes 150 CMR short-axis stacks featuring end-diastolic and end-systolic frames acquired from equal numbers of subjects from 5 groups: patients with 4 different cardiac pathologies and healthy subjects\footnote{For more details see: \url{https://acdc.creatis.insa-lyon.fr/}}. The data are split into 100 subjects for training and 50 for testing. In this experiment our task was to segment the myocardium of the left ventricle. We applied our method using the postprocessing framework, since the topologies of the myocardium in different slices of the short-axis stack can vary. We first trained a U-net model \cite{Ronneberger2015} using a binary cross entropy loss on the 100 training cases. Next, the topological prior knowledge for the 50 test cases was specified by manual inspection of all slices. Three different types of topology were present in the test set: a single connected component with one loop (i.e. $\beta_0^* = \beta_1^*=1$), a single connected component with no loops ($\beta_0^*=1$, $\beta_1^*=0$) and no connected components ($\beta_0^*=0$, $\beta_1^*=0$). This prior knowledge was used when applying the trained model using the postprocessing framework with $\lambda = 0.01$ (see Equation \ref{eq:post_proc_framework}).

We achieved mean Dice scores of 0.8994 and 0.9068 at end-diastole and end-systole respectively. These results are comparable with state-of-the-art techniques and within the range of agreement of the leading method reported in \cite{Bernard2018}. Note that we are unable to assess topological correctness for this experiment due to the lack of public availability of the ground truth segmentations for the test set.\\

\subsection{Experiment 4}
\label{sec:experiment_4}
\begin{figure}[!t]
\begin{subfigure}{3.4in}
\centering
\includegraphics[width=1.7in]{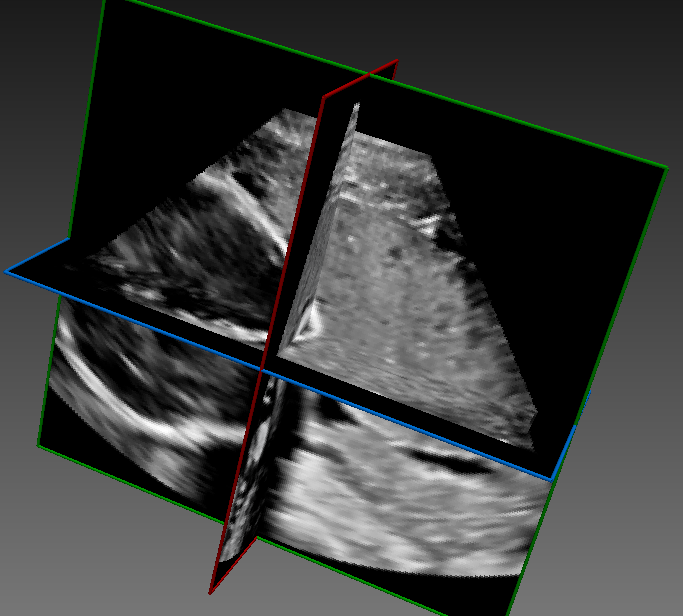}%
\includegraphics[width=1.7in]{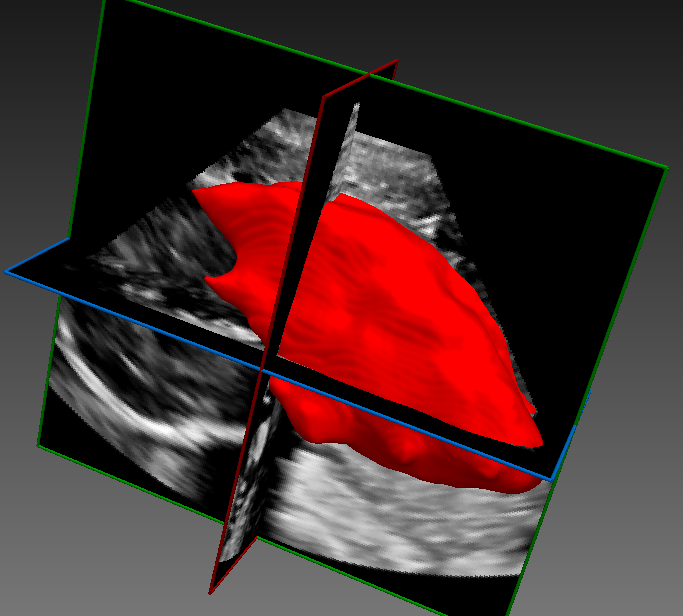}
\caption{Left: an example 3D ultrasound volume.
Right: the ground-truth segmentation in red.}
\end{subfigure}
\begin{subfigure}{3.4in}
\centering
\includegraphics[width=1.7in]{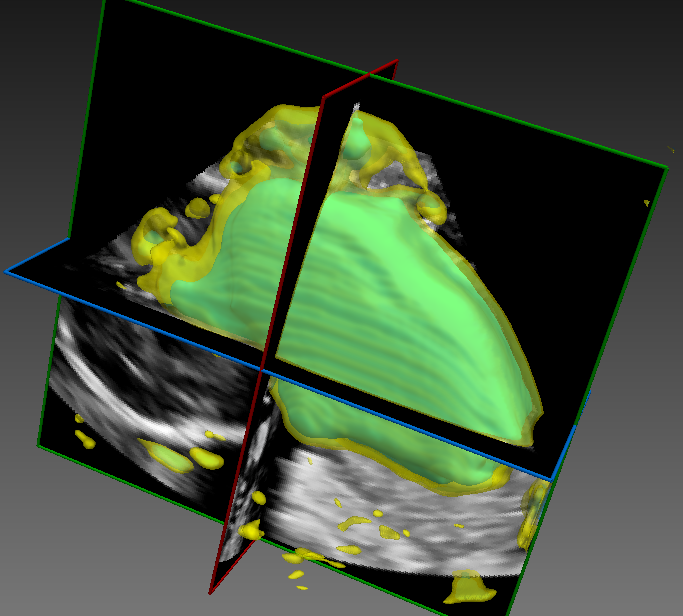}%
\includegraphics[width=1.7in]{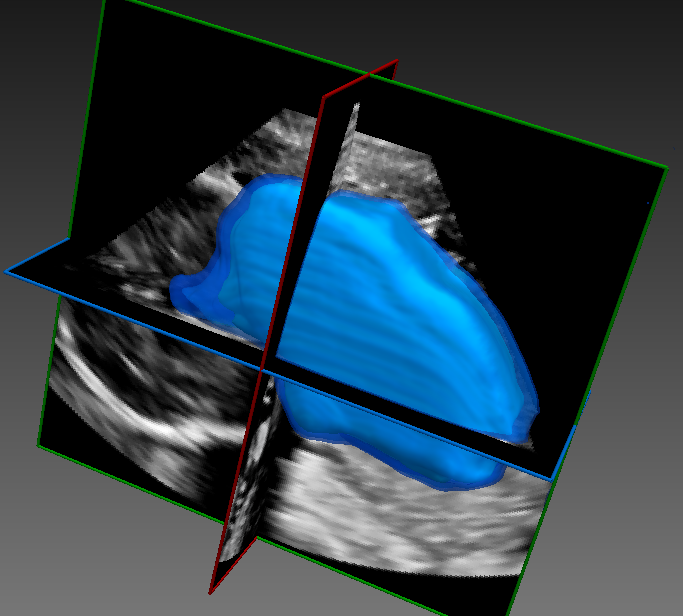}
\caption{Left: yellow-green contours showing the predicted segmentation from the 3D U-net trained only with the Dice loss.
Right: blue contours showing the predicted segmentation from the 3D U-net trained with the Dice and topological loss.}
\end{subfigure}
\begin{subfigure}{3.4in}
\centering
\includegraphics[width=1.7in]{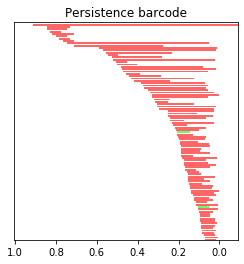}%
\includegraphics[width=1.7in]{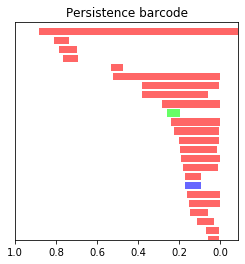}
\caption{Left: the persistence barcode diagram of the left segmentation in (b).
Right: the persistence barcode diagram of the right segmentation in (b).}
\end{subfigure}
\caption{The segmentation provided by the network trained with only the Dice loss has several connected components outside of the main segmented object.
Its persistence barcode contains many bars demonstrating that there are a large number of disconnected components segmented throughout the volume.
The topological loss function encourages the network to output a segmentation with fewer connected components, as can be seen from the output segmentation on the right.
The corresponding persistence barcode diagram has many fewer bars, demonstrating that this output is closer, in a topological sense, to the ground truth segmentation as well as having a higher Dice coefficient with the ground truth.}
\label{fig:placenta_topoprior_example}
\end{figure}

To demonstrate the applicability of our method to 3D volumes, as well as to other imaging modalities we performed a final experiment in which the task was to segment the placenta in 3D ultrasound volumes of pregnant women.
$17$ patients in the third trimester ($29-34$ weeks of gestation) were scanned using a Philips EPIQ 7g and a x6-1 transducer. 
The 3D ultrasound volumes were selected from 4D (3D+t) image streams covering different parts of the placenta. 
The annotations were produced manually by an expert sonographer. 
In total $67$ annotated volumes were used in this experiment.
Volumes were chosen such that those coming from the same patient covered different regions of the placenta and so were not too similar to each other.
Each volume was cropped to 96x240x256 voxels and voxel intensities normalised to the $[0,1]$ range. 
Figure \ref{fig:placenta_topoprior_example} shows a typical case.

\begin{figure}[!t]
    \centering
    \includegraphics[width=3in]{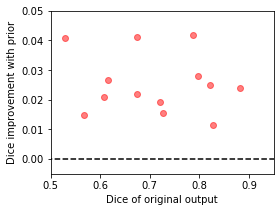}
    \caption{The improvement in the Dice scores of the segmentations of the placenta when using the topological prior post-processing.
    We split the dataset into 13 folds, training the network on 12 and testing on the other 1.
    Each point here is the average for each of the folds, which contain 5 or 6 volumes each.
    The difficulty of the segmentation task varies significantly between volumes causing the wide range of Dice scores between folds.
    Nonetheless, applying the topological prior in post-processing consistently improves the resulting segmentations by an average of $0.024$.}
    \label{fig:placenta_dice_improvement}
\end{figure}

In this experiment the segmentation network is a 3D U-net \cite{cciccek20163d} and the relevant topological prior knowledge is that the segmented placenta should form one single connected component and there should be no loops/handles or cavities within it, i.e. that $\beta_0^*=1$ and $\beta_1^*=\beta_2^*=0$.
As the total number of cases we have available is limited, we use our method in the post-processing framework, minimising the loss in Equation \ref{eq:post_proc_framework}.
Figure \ref{fig:placenta_dice_improvement} shows that applying the topological prior in post-processing, with $\lambda = 0.005$ results in a consistent improvement in the Dice score of the segmentations, with an average improvement (across a 13-fold cross-validation) of $0.024$.
Figure \ref{fig:placenta_topoprior_example} shows a typical case without, and with the topological prior.
Introducing this prior dramatically reduces the number of small components in the segmentation as well as loops and cavities in the large component.

\section{Discussion}
The key contribution of this work is to demonstrate that PH is a viable tool for adding a topological loss function to train CNNs to perform image and volume segmentation.
We have shown that by using a U-net-like network architecture and supplementing traditional pixel-wise loss functions with our topological loss function, the accuracy of the resulting segmentations can be improved in terms of the pixel-wise accuracy and also that there are significant improvements in terms of their topological accuracy.
Of course there will  be limits to the improvements that the topological loss can make. 
If the predicted segmentation is already topologically correct then it will make little or no difference. 
Conversely, if the predicted segmentation is too far from the ground truth it may not be possible to recover the correct topology as gradient descent of the topological loss function will not necessarily reach the global minimum. 
As a simple intuitive example, if the output is expected to contain one connected component and the predicted segmentation has more than one large component, the network will not know which to encourage and which to suppress. 
In practice we have found that the `basin of attraction' for our topological loss function is large in that such failure cases are rare. Indeed, in Experiment 2 we demonstrated that our PH based method outperformed a range of comparative techniques at corruption levels above what we would expect to encounter in realistic clinical scenarios. In order to be able to confidently apply our method in different domains, we would advise a similar analysis of robustness to noise/corruption levels to be carried out.

Although PH has been used in a wide variety of applications we believe that medical image analysis is a particularly appealing one.
This is because machine learning tasks in medical imaging often deal with small datasets (due to the expense of acquiring data, and privacy concerns with sharing them) and hard to interpret or noisy images (due to motion artefacts or the desire to acquire images quickly).
But they also often come equipped with highly informative prior knowledge, since we know which anatomy is being imaged, its approximate location in the image and the parameters and protocols of the image acquisition. 
To be able to make use of this prior knowledge we need to be able to integrate it with powerful statistical methods such as deep neural networks, and PH is a strong candidate for bridging this gap.
 In general, our method is applicable in cases where limited training data are available, but prior knowledge of topological properties is available \emph{a priori}. We have demonstrated a number of such situations in medical imaging. 
Beyond medical imaging, we believe that our method could potentially be beneficial in other tasks such as segmenting video images of pavements and aerial images of roads \cite{Mosinska2018} or astronomy, in all of which it is likely that the topology of the structures being segmented would be known.

In \cite{Clough2019} we presented our preliminary work.
The most significant difference between our earlier method and this presented work is that here we explicitly define a loss function based on PH rather than using PH to derive a gradient used in training.
Our previous method had the drawback that it was difficult to know whether or not gradient descent for the derived gradient converged and how many steps this would take.
Conversely, with the method presented here, the scalar loss function allows the training progress to be observed easily and to stop training when the validation loss is minimised.
By avoiding the iterative process described in algorithm 1 in \cite{Clough2019} we do not have to choose the hyperparameter defining the number of pixels to identify in that iterative process.
The fact that the loss used here is proportional to the squared length of the bars in the barcode diagram means that longer unwanted bars are naturally penalised more than shorter ones, removing the need for thresholding on bar length, and resulting in quicker training.
Finally, in this work we demonstrate the viability of this approach on 3D volumes, on imaging modalities beyond CMR, and more than just one topological prior.

Calculating the persistent homology of each candidate segmentation adds a computational cost to training the network.
The PH for a cubical complex of dimension $d$ and with $V$ pixels/voxels can be computed in $\Theta(3^d V + d2^dV)$ time and $\Theta(d2^dV)$ memory \cite{Wagner2012}.
It is worth noting here that we require not just the birth/death thresholds for each feature but also their gradients with respect to the input object.
In our experiments we found that calculating the PH on a batch of 100 images of size 80x80 took approximately 10 seconds.
The calculation for 1 volume of size 96x240x256 took approximately 6 seconds.
Whether this additional calculation time is acceptable or not depends upon the application in question.
Using 3D segmentations to print patient-specific models of anatomy is already a time-consuming process and so adding seconds or even several minutes to the time taken to perform segmentation is acceptable.
Where segmentations are required in real-time, the computational cost of applying our method in the post-processing framework may become prohibitive, at least in 3D.
However the PH calculation is open to optimisation schemes in that it can be much more quickly calculated on downsampled versions of the proposed segmentation. Computational optimisation is not the focus of this paper but we believe that it may be possible to improve efficiency to the extent that it becomes acceptable for many applications.

Aside from the cardiac and placental segmentation problems demonstrated here, we believe that our approach will be applicable to other tasks in which topology is relevant in segmentation.
Many neuroimaging pipelines begin with the segmentation of the cortical surface from MR volumes of the brain.
In order to compare cortical surfaces the segmented region must be a topological sphere.
Current standard approaches involve retrospective topology correction \cite{Segonne2007} of the segmentation in order to ensure this.
Our approach would consider the need for a topologically correct surface to be segmented to be an inherent part of the segmentation task itself.
Similarly, vascular tree segmentation \cite{fraz2012blood} is a case where post-processing for topology correction could be replaced with topological priors inside the network performing segmentation.

In some applications, topological accuracy can be more important than pixelwise accuracy.
The ability to trade off the two with a weighting parameter in the loss function is a benefit of our approach.
An example is in the segmentation of CMR volumes of patients with congenital heart defects for the purpose of patient-specific 3D printing \cite{vukicevic2017cardiac, byrne2016systematic}.
In this application, segmenting the septal walls of the atria and ventricles with the correct thicknesses is often not vital since the printing process places its own constraints on these parameters.
What is important though is correctly segmenting the holes between the chambers as these abnormal connections are the details that are important to the surgeon using the 3D model \cite{Byrne2019}. 
Therefore, topological accuracy is more relevant to the task at hand than, for example, the Dice coefficient, and so a user could adjust the loss function when training a network to perform the segmentation task to reflect this.
A further possible application-relevant generalisation of our approach is multi-class segmentation tasks, in which the topology of each class and also of the boundaries between each class could be specified. This can be thought of as applying PH to the problem tackled in \cite{Ganaye2018} in which the adjacencies of various brain regions are specified as a prior.

One of the benefits of our topological loss function is that it can suppress small false positive or false negative regions in the predicted segmentation (because these would change the topology of the segmentation). Other techniques exist to suppress such regions, such as morphological operations or CRF based techniques \cite{Krahenbuhl2011,Zheng2015}. However, whilst these approaches can correct such local errors, they do not have any notion of global topology, only local label smoothness. The advantage of our PH-based approach is that the correct global topology can be encouraged, whether or not this also encourages label smoothness. For example, two large components (such as chambers of the heart) may be encouraged to join together using a small connecting region (such as a structural defect) if the expected global topology specifies that they should be a single component rather than two separate components.

While our method attempts to add prior knowledge to segmentation networks by creating a loss function which measures the degree to which the network's output adheres to the prior, an alternative approach is to begin with a shape model which has the desired shape and/or topology and then to learn a deformation which fits that model to the data.
While shape models have a long history in medical image analysis \cite{pizer2003deformable}  traditional methods require solving an optimisation problem at inference time.
More recent work such as \cite{Chung2019} attempts to train a neural network to perform the deformation required to fit a pre-defined shape model to an image, such that at inference time only one forward pass through the network is required and so inference takes milliseconds rather than minutes or hours.
Nonetheless there is a fundamental limitation of all deformable model methods, which is that there are limits on how much the initial shape model can deform.
This is  particularly important when segmenting unusually shaped but topologically correct anatomy where large deformations would be required.

\section{Conclusion}
We have presented a loss function for training CNNs to perform image segmentation which assesses the extent to which the proposed segmentation adheres to our prior knowledge of its topology.
Using persistent homology, the robustness of various topological features can be computed in a manner which allows for gradient descent on the weights of the network performing the segmentation.
Our experiments have shown that our approach is applicable to 2D images and 3D volumes, to CMR imaging and ultrasound, and that in these cases it improves the pixelwise and topological accuracy of the resulting segmentations.


%

\ifCLASSOPTIONcompsoc
  \section*{Acknowledgments}
\else
  \section*{Acknowledgment}
\fi
This work was supported by 
the EPSRC programme Grant `SmartHeart' (EP/P001009/1), 
the Wellcome Trust IEH Award [102431], 
the Wellcome EPSRC Centre for Medical Engineering at School of Biomedical Engineering and Imaging Sciences, King�s College London (WT 203148/Z/16/Z) 
and by the National Institute for Health Research (NIHR) Biomedical Research Centre at Guy�s and St Thomas� NHS Foundation Trust and King�s College London.
This research has been conducted using the UK Biobank Resource under Application
Numbers 17806 and 40119. 
We would like to thank Nvidia for kindly donating the Quadro P6000 GPU used in this research.

\section*{Code/Data Availability}

The code for running the MNIST experiment is available from: \url{https://github.com/JamesClough/topograd}. The MNIST dataset
is available from: \url{http://yann.lecun.com/exdb/mnist/}. The
UK Biobank dataset is available to approved projects from: \url{https://www.ukbiobank.ac.uk/}. The ACDC dataset is available from: \url{https://acdc.creatis.insa-lyon.fr/}. We are unable to release the placenta dataset due to lack of patient consent for data sharing.

\ifCLASSOPTIONcaptionsoff
  \newpage
\fi



\bibliographystyle{IEEEtran}
\bibliography{IEEEabrv,refs}
%

%

\begin{IEEEbiography}[{\includegraphics[width=1in,height=1.25in,clip,keepaspectratio]{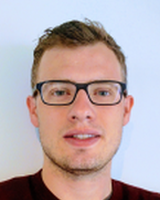}}]{James R Clough}
received an MSci degree in
Theoretical Physics and a PhD degree in Physics from Imperial College London, in 2013 and
2017, respectively. He was a Research Associate at
King's College London in the Biomedical Engineering Department. His research focused on
machine learning, particularly as applied to medical image analysis, multi-modal data and graphs.
\end{IEEEbiography}%
\begin{IEEEbiography}[{\includegraphics[width=1in,height=1.25in,clip,keepaspectratio]{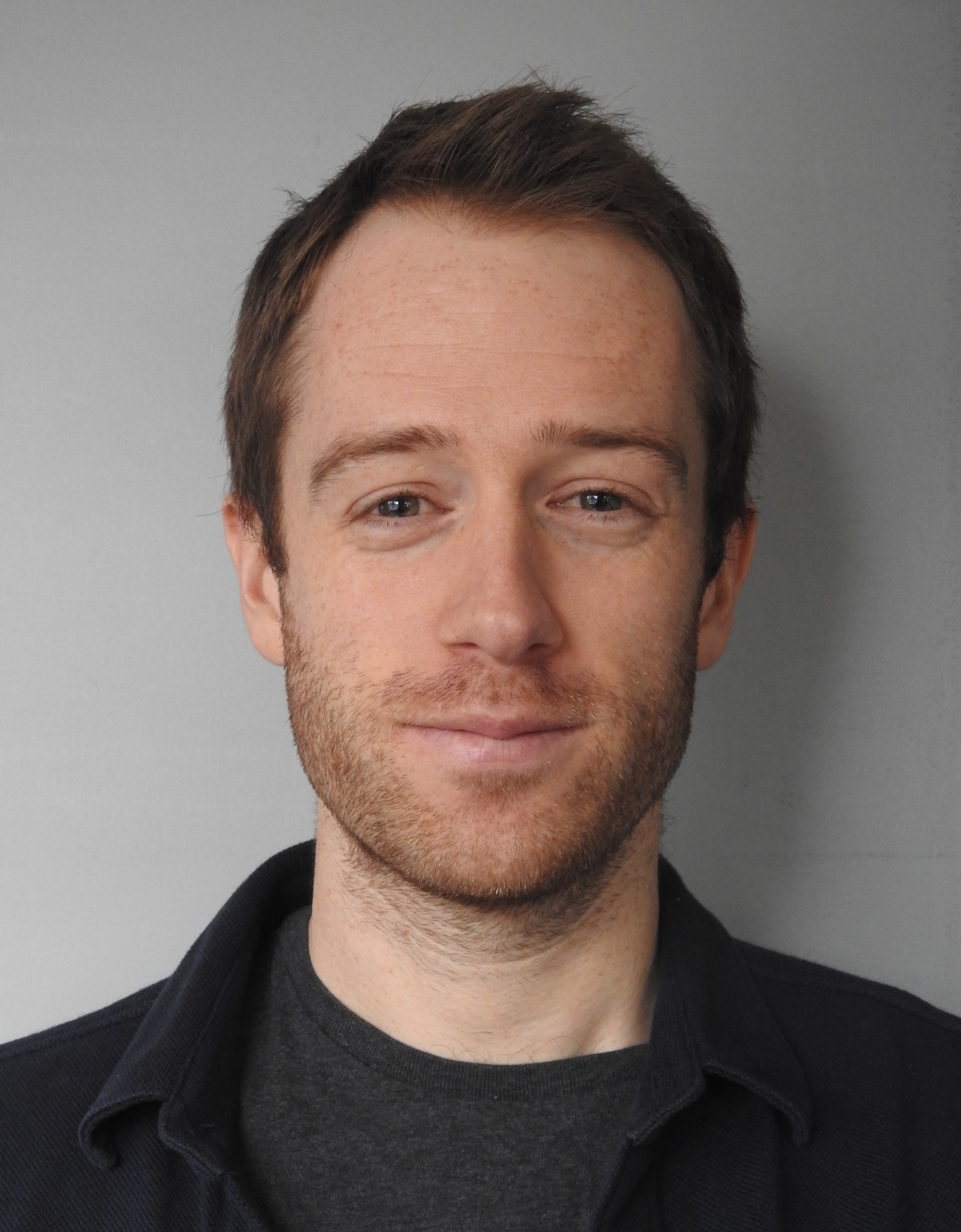}}]{Nicholas Byrne}
received an MPhys degree in Physics from the University of Warwick, and an MSc in Clinical Sciences (Medical Physics) and MRes in Clinical Research, both from King's College London.
He is currently a registered medical physicist at Guy's and St. Thomas' NHS Foundation Trust and a PhD student at King's College London.
He is interested in methods for the processing and visualisation of 3D medical images, and in particular, patient-specific 3D printing for patients with congenital heart disease.
\end{IEEEbiography}%
\begin{IEEEbiography}[{\includegraphics[width=1in,height=1.25in,clip,keepaspectratio]{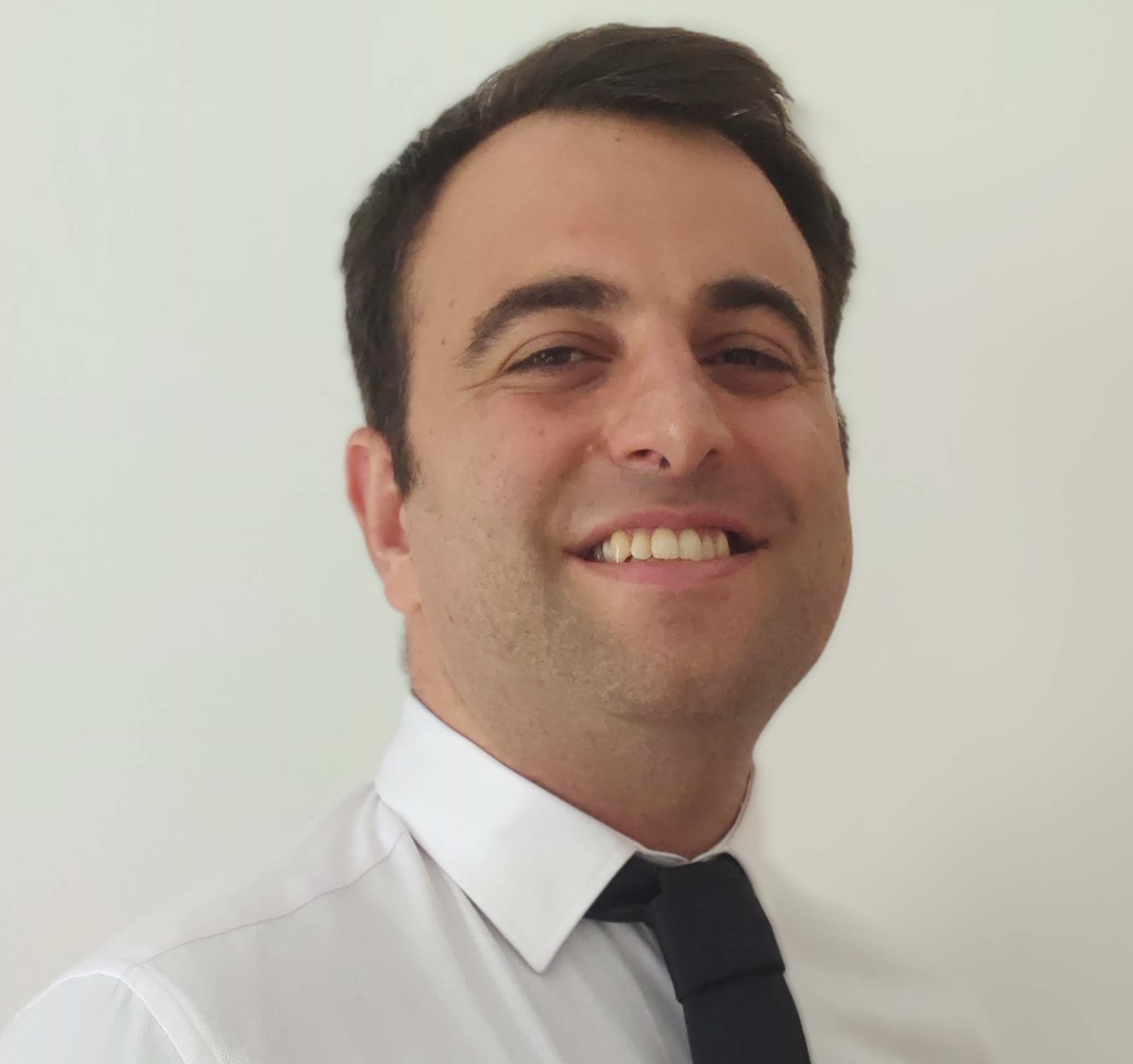}}]{Ilkay Oksuz}
received  his BSc degree from Istanbul Technical University in Electronics Engineering and his MSc degree from Bahcesehir University in Electrical and Electronics Engineering. He received a PhD degree from IMT School for Advanced Studies Lucca on Computer, Decision, and Systems Science. He was a Research Associate at
King's College London in the Biomedical Engineering Department from 2017 until 2019. He has been working as an Assistant Professor at Istanbul Technical University in the Computer Engineering Department since 2020. His work focuses on machine learning with a particular interest in medical image quality assessment,  reconstruction and analysis.
\end{IEEEbiography}%
\begin{IEEEbiography}[{\includegraphics[width=1in,height=1.25in,clip,keepaspectratio]{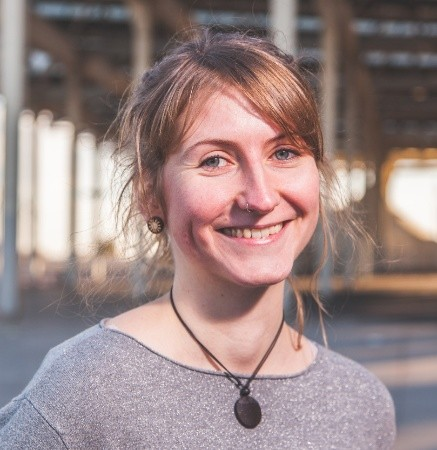}}]{Veronika A. Zimmer} received her BSc and MSc degrees in Computational Life Science from the University of L\"{u}beck, Germany, in 2008 and 2011, respectively.  She received her PhD in Information and Communication Technologies from the Universitat Pompeu Fabra, Barcelona, Spain, in 2017. She is a Research Fellow at King's College London in the School of Biomedical Engineering and Imaging Sciences. Her research focuses on image analysis and machine learning, in particular multimodal registration and segmentation in medical imaging.

\end{IEEEbiography}%
\begin{IEEEbiography}[{\includegraphics[width=1in,height=1.25in,clip,keepaspectratio]{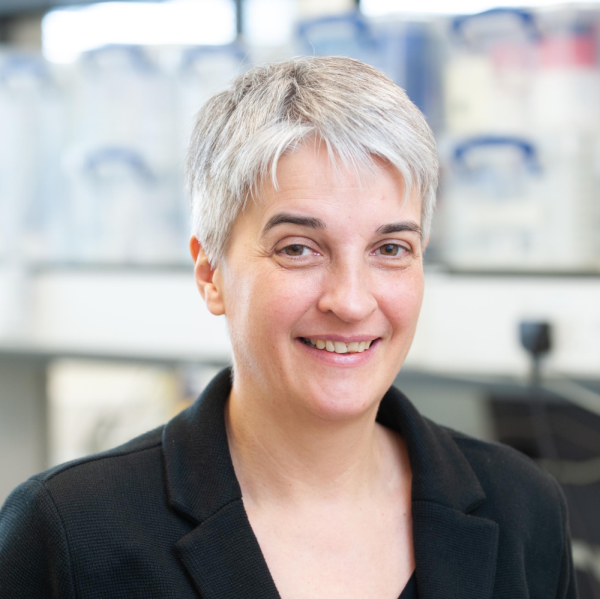}}]{Julia A. Schnabel}
received a Diploma in Computer Science from the Technical University Berlin and a PhD in Computer Science from University College London. She was a post-doctoral research associate/fellow at University Medical Center Utrecht, King's College London and University College London, before joining the University of Oxford in 2007 as Associate Professor in Engineering Science (Medical Imaging), and Fellow of Engineering at St. Hilda's College, Oxford. In 2014 she became Professor of Engineering Science by Recognition of Distinction, and in 2015 she joined the School of Biomedical Engineering and Imaging Sciences at King's College London as Chair in Computational Imaging. Her research focuses on machine/deep learning, nonlinear motion modelling, as well as multi-modality, dynamic and quantitative imaging for a wide range of medical imaging applications.
\end{IEEEbiography}%
\begin{IEEEbiography}[{\includegraphics[width=1in,height=1.25in,clip,keepaspectratio]{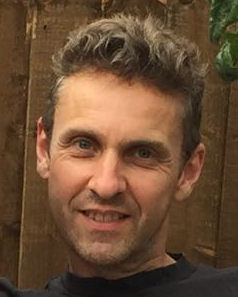}}]{Andrew P. King}
received a BSc degree in
Computer Science from Manchester University,
an MSc degree in Cognition, Computing and Psychology from Warwick University, and a PhD
degree in Computer Science from Warwick University. He is a Reader in Medical Image Analysis
in the Biomedical Engineering Department,
King's College London. From 2001-2005 he
worked as an Assistant Professor in the Computer
Science Department, Mekelle University in Northern Ethiopia. Since 2006 he has worked in the
Biomedical Engineering Department, King�s College London, focusing
on image analysis and machine learning in medical imaging.
\end{IEEEbiography}%






\end{document}